 \documentclass[final,5p,times,twocolumn]{elsarticle}

\usepackage{amssymb}
\usepackage{lipsum}
\usepackage{amsmath,amssymb,amsfonts}
\usepackage{algorithmic}
\usepackage{graphicx}
\usepackage{textcomp}
\usepackage{xcolor}
\usepackage{float} 
 \usepackage{epstopdf}
 \usepackage{multirow}
\usepackage{makecell}
\usepackage[caption=false,font=footnotesize,labelfont=rm,textfont=rm]{subfig}
\DeclareUnicodeCharacter{230A}{\lfloor}
\usepackage{booktabs}

\journal{Nuclear Physics B}

\begin{document}

\begin{frontmatter}



\title{A Privacy-Preserving Framework  with Multi-Modal Data for Cross-Domain Recommendation}






\author[first]{Li Wang}
\ead{li.wang-13@student.uts.edu.au}
\author[second]{Lei Sang}
\ead{ sanglei@ahu.edu.cn}
\author[third]{Quangui\ Zhang}
\ead{zhqgui@126.com}
\author[first]{Qiang Wu}
\ead{Qiang.Wu@uts.edu.au}
\author[first]{Min Xu\corref{cor1}}
\ead{Min.Xu@uts.edu.au}
\affiliation[first]{organization={School of Electrical and Data Engineering,University of Technology Sydney},
            addressline={15, Broadway, Ultimo}, 
            city={Sydney},
            postcode={2000}, 
            state={NSW},
            country={Australia}}
\affiliation[second]{organization={School of Computer Science and Technology, Anhui University},
            addressline={111, Jiulong Road, Economic and Technological Development District}, 
            city={Hefei},
            postcode={230601}, 
            country={China}}
\affiliation[third]{organization={School of Artificial Intelligence,
Chongqing University of Arts and Sciences},
            addressline={319, Honghe Avenue, Yongchuan District}, 
            city={Chongqing},
            postcode={402160}, 
            country={China}}
\cortext[cor1]{Corresponding author}

\begin{abstract}
Cross-domain recommendation (CDR) aims to enhance recommendation accuracy in a target domain with sparse data by leveraging rich information in a source domain, thereby addressing the data-sparsity problem.
Some existing CDR methods highlight the advantages of extracting domain-common and domain-specific features to learn comprehensive user and item representations.
However, these methods can't effectively disentangle these components as they often rely on simple user-item historical interaction information (such as ratings, clicks, and browsing), neglecting the rich multi-modal features.
Additionally, they don't protect user-sensitive data from potential leakage during knowledge transfer between domains.
To address these challenges, we propose a \textbf{P}rivacy-\textbf{P}reserving Framework with \textbf{M}ulti-\textbf{M}odal Data
for \textbf{C}ross-\textbf{D}omain \textbf{R}ecommendation, called P2M2-CDR.
Specifically, we first design a multi-modal disentangled encoder that utilizes multi-modal information to disentangle more informative domain-common and domain-specific embeddings.
Furthermore, we introduce a privacy-preserving decoder to mitigate user privacy leakage during knowledge transfer.
Local differential privacy (LDP) is utilized to obfuscate the disentangled embeddings before inter-domain exchange, thereby enhancing privacy protection.
To ensure both consistency and differentiation among these obfuscated disentangled embeddings, we incorporate contrastive learning-based domain-inter and domain-intra losses.
Extensive Experiments conducted on four real-world datasets demonstrate that P2M2-CDR outperforms other state-of-the-art single-domain and cross-domain 
baselines. 
\end{abstract}



\begin{keyword}
Privacy-Preserving \sep Multi-Modal \sep Disentanglement \sep Contrastive Learning \sep Cross-Domain Recommender Systems
\end{keyword}

\end{frontmatter}


\section{Introduction}

Cross-domain recommendation (CDR) emerges as a crucial strategy to tackle the enduring issue of data sparsity in recommendation systems by transferring informative knowledge across related domains \cite{zhu2021cross,hu2018conet}. 
The inherent data-sparsity problem arises when the target domain lacks sufficient user-item interaction information, hindering the ability to provide accurate and personalized recommendations. 
To address this challenge, various methods \cite{hu2018conet,xin2015cross,zhu2020deep,man2017cross, zhao2020catn,li2020ddtcdr,zhu2019dtcdr,liu2020cross,zhu2020graphical} have been devised to enhance the performance of cross-domain recommendation systems.
These methods leverage the rich knowledge in the source domain to complement the sparse information in the target domain.
Collective Matrix Factorization (CMF)-based methods \cite{xin2015cross,zhu2020deep} concentrate on creating shared user and item representations in different domains by matrix factorization technology.
In contrast, embedding-based methods \cite{man2017cross,zhao2020catn} train separate encoders to learn embeddings and subsequently employ mapping techniques to project user and item embeddings into a shared space.
Dual knowledge transfer-based methods \cite{hu2018conet,li2020ddtcdr,zhu2019dtcdr,liu2020cross} are dedicated to bidirectional knowledge transfer, while Graph Neural Networks (GNNs)-based methods \cite{zhu2020graphical} leverage high-order collaborative information to enhance performance.
However, these approaches often assume that users in different domains have the same interests.
They tend to learn user representations by relying on shared knowledge, thereby overlooking the possibility that users may have diverse preferences across different domains.
This means that they focus solely on domain-common information while ignoring domain-specific information.

\begin{figure*}[!htb]
\centering
\subfloat[]{\includegraphics[width=2.9in]{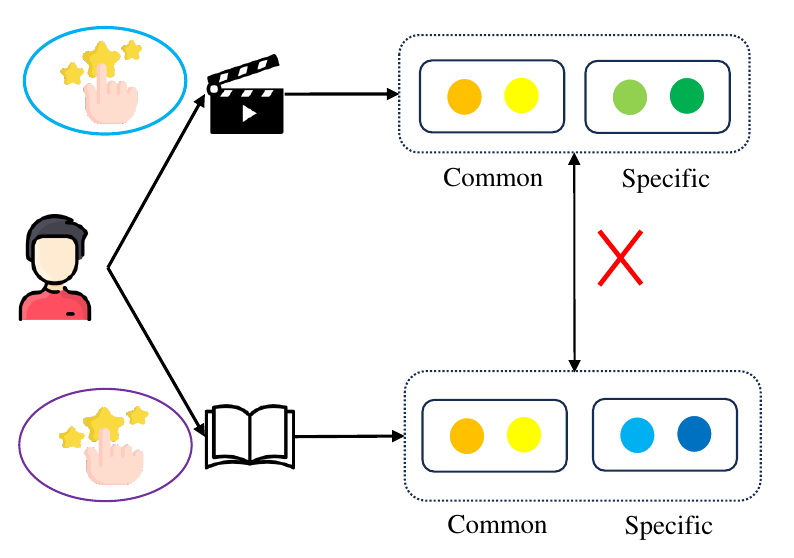}%
\label{existing-CDR}}
\hfil
\subfloat[]{\includegraphics[width=3.6in]{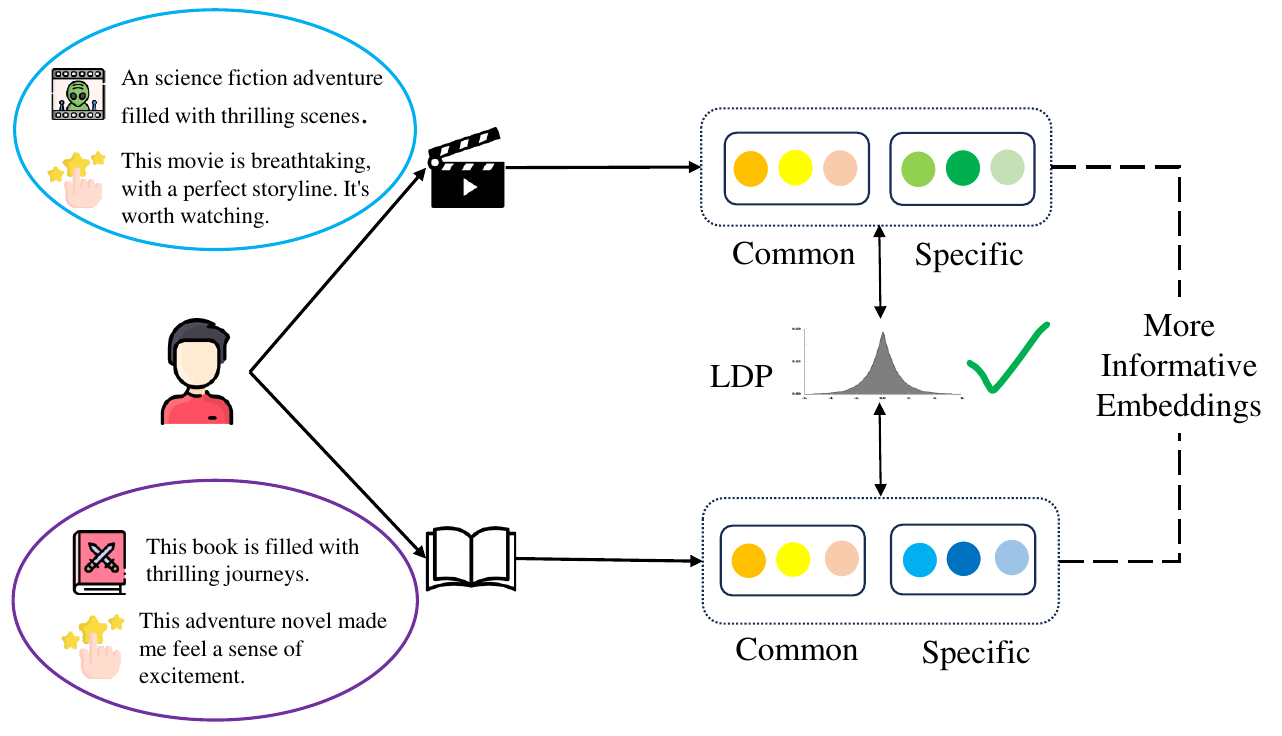}%
\label{P2M2-CDR-motivation}}
\caption{An illustration of earlier CDR methods (a) and P2M2-CDR (b). In comparison to earlier CDR methods, P2M2-CDR considers (1) introducing multi-modal data (review texts, textual, and visual features) to disentangle more informative domain-common and domain-specific features and (2) utilizing local differential privacy technology to protect user privacy.}
\label{motivation}
\end{figure*}

It's crucial to separate domain-common and domain-specific knowledge to make them contain more semantic information.
In recent years, several methods \cite{liu2020cross, cao2022disencdr, zhang2023disentangled, chen2019efficient} have emerged with the aim of simultaneously capturing both domain-common and domain-specific knowledge across different domains.
As depicted in Figure \ref{existing-CDR}, these methods typically employ user-item interaction history to design methods that disentangle domain-common and domain-specific features.
Subsequently, these features are aggregated to facilitate the learning of user and item representations.
For example, DisenCDR \cite{cao2022disencdr} introduces two mutual information-based regularizers to effectively disentangle domain-common and domain-specific components.
Similarly, DCCDR \cite{zhang2023disentangled} utilizes contrastive learning to decouple these two components.
Additionally, DIDA-CDR \cite{zhu2023domain} employs domain classifiers to disentangle domain-common, domain-specific, and domain-independent features, aiming to learn more comprehensive user representations.
While these methods have demonstrated improved performance, they encounter two limitations.

\noindent (1) \textbf{Limitation in Domain Disentanglement.} They solely depend on user-item historical interaction information (such as ratings, clicks, and browsing) to extract domain-common and domain-specific features.
However, the recommendation accuracy may decline because only utilizing this simple user-item interaction data can't adequately decouple these features.

\noindent (2) \textbf{Limitation in Privacy Protection.} These CDR models assume that the disentangled embeddings between different domains are public and shared.
However, they do not address the user privacy problem during knowledge transfer between domains.
Since these disentangled embeddings are linked to sensitive user data, such as user-item ratings, sharing them increases the risk for external participants and potential attackers to infer user privacy information.
Moreover, in real-world scenarios, these domains might be operated by different organizations. Sharing these embeddings could potentially violate business privacy policies.

Based on the above limitations, in this paper, we primarily address two challenges in designing a privacy-preserving cross-domain recommendation system.
\textbf{CH1.} How can we effectively decouple more informative domain-common and domain-specific embeddings to learn comprehensive user and item representations?
\textbf{CH2.} How do we prevent the leakage of user-related information when learning and transferring disentangled embeddings?
To tackle these challenges, we introduce P2M2-CDR, a privacy-preserving framework for cross-domain recommendation, as illustrated in Figure \ref{P2M2-CDR-motivation}.
This framework leverages multi-modal data to efficiently disentangle domain-common and domain-specific features.
Additionally, we incorporate local differential privacy (LDP) [15] to mitigate the risk of user privacy leakage during knowledge transfer.

To address the first challenge, we propose a multi-modal disentangled encoder that incorporates multi-modal information to effectively disentangle more informative domain-common and domain-specific embeddings.
Multi-modal information provides a more comprehensive and richer description of users and items \cite{mcauley2015image, cheng2018aspect}.
By integrating these diverse sources, the model gains a deeper understanding of users' behaviors and preferences.
Consequently, it becomes more proficient at distinguishing between common and specific characteristics of users and items across different domains.
Specifically, for both the source and target domains, we initially learn item ID embeddings from LightGCN \cite{he2020lightgcn}, a model that captures high-order collaborative relationships in the user-item interaction graph. We then combine these item ID embeddings with item textual features obtained from the pre-trained Sentence-Transformer model \cite{reimers2019sentence} and item visual features extracted from VGG19 \cite{simonyan2014very} to obtain item representations. 
Simultaneously, we incorporate user ID embeddings learned from LightGCN and user review text embeddings acquired from Sentence-Transformer to effectively decouple domain-common and domain-specific embeddings. Moreover, we employ self-supervised techniques, such as the feature dropout strategy, to generate augmented disentangled embeddings.

To address the second challenge, we propose a privacy-preserving decoder that leverages local differential privacy techniques to safeguard user-related data during the transfer of disentangled embeddings.
Initially, we employ local differential privacy (LDP) to obscure domain-common and domain-specific embeddings before inter-domain sharing, thereby safeguarding user-privacy information from participants and external threats. 
Subsequently, to ensure the alignment and separation of these obfuscated decoupled embeddings, we design contrastive learning-based domain-inter and domain-intra losses.
Finally, we aggregate the obfuscated domain-common and domain-specific embeddings to learn comprehensive user preferences.
 
In summary, our proposed model makes the following contributions:
\begin{itemize}
\item We design a multi-modal disentangled encoder that utilizes multi-modal information, including user review texts, item textual features, and item visual features, to disentangle more informative domain-common and domain-specific embeddings (for \textbf{CH1}).
\item We propose a privacy-preserving decoder that employs local differential privacy techniques to prevent the leakage of user privacy information during the transfer of disentangled embeddings (for \textbf{CH2}).
\item We conducted extensive experiments on four large-scale real-world datasets from Amazon. Comprehensive results demonstrate the effectiveness of P2M2-CDR compared with some state-of-the-art baselines.
 \end{itemize}

 \section{Related Work}
\subsection{Cross-Domain Recommendation}
Cross-domain recommendation system (CDR) aims to address the challenges of information sharing and knowledge transfer across diverse domains or platforms, providing more accurate and personalized recommendation services. CDR is generally categorized into three types: single-target recommendation \cite{zhao2020catn, fu2019deeply}, dual-target recommendation \cite{cao2022disencdr, zhu2023domain}, and multi-target recommendation \cite{guo2023disentangled, mukande2022heterogeneous}. The core step in CDR involves designing an effective transfer method capable of transferring relevant knowledge from a rich domain to a sparse domain, thereby enhancing recommendation accuracy.
With the advancements in deep learning, various transfer methods have emerged in CDR, including domain transfer techniques \cite{yu2020semi}, learning mapping functions across domains \cite{man2017cross}, deep dual knowledge transfer \cite{li2020ddtcdr}, and transfer methods based on Graph Neural Networks (GNNs) \cite{cui2020herograph, zhao2019cross}.
However, these approaches often assume that data across all domains is openly shared, potentially overlooking user privacy concerns.
In this paper, our primary focus is on mitigating the risk of user privacy leakage through the utilization of local differential privacy techniques.
\subsection{Disentangled Representation Learning in Recommendation}
Disentangled representation learning in recommendation aims to decompose the user's characteristics or item features into distinct and independent parts, providing an effective way to enhance the robustness and interpretability of models.
Disentangled representation learning has been applied in generative recommendations \cite{ma2019learning}, causal recommendations \cite{zheng2021disentangling}, and graph recommendations \cite{wang2020disentangled}.
Authors in \cite{ma2019learning} propose a model called MacridVAE, which is a disentangled variational auto-encoder achieving both macro disentanglement of high-level concepts and micro disentanglement of isolated low-level factors. 
DICE \cite{zheng2021disentangling} constructs cause-specific data based on causal effects and disentangles user and item embeddings into interest and conformity components.
DGCF \cite{wang2020disentangled} learns disentangled representations that capture fine-grained user intents from the user-item interaction graph.

Recently, disentangled representation learning has been applied to cross-domain recommendation systems.
For example, DisenCDR \cite{cao2022disencdr} disentangles user preferences into domain-specific and domain-shared information with two mutual information-based regularizers 
and transfers the domain-shared information across both domains. Zhu et al. \cite{zhu2023domain} propose a disentanglement framework to decouple domain-specific, domain-independent, and domain-shared information for learning comprehensive user preferences.
However, these methods often neglect multi-modal information when disentangling domain-common and domain-specific features, resulting in suboptimal recommendation performance.
In our approach, we effectively leverage multi-modal features to overcome this limitation.
\subsection{Privacy-Preserving Recommendation}
Existing privacy-preserving recommendation systems \cite{wu2022fedcl, wei2022heterogeneous,wu2021fedgnn,ammad2019federated,yu2021privacy,gao2019cross} are designed to offer personalized recommendations while ensuring the protection of user privacy.
Nowadays, various privacy protection technologies, such as local differential privacy and federated learning, have gained popularity in CDR \cite{gao2019privacy, yan2022fedcdr, chen2022differential, chen2023win}.
These technologies ensure that users’ sensitive information is not misused or disclosed during the knowledge transfer process.
For instance, in \cite{yan2022fedcdr}, the authors propose FedCDR, a privacy-preserving federated CDR framework.
This framework learns user and item embeddings through individual models trained on personal devices.
Following this, it uploads the weights to the central server to protect the user's sensitive data. 
Other approaches like PriCDR \cite{chen2022differential}, CCMF \cite{gao2019privacy} and PPGenCDR \cite{Liao_Liu_Zheng_Yao_Chen_2023} employ differential privacy techniques to construct a protected rating matrix in the source domain. 
Subsequently, they transfer the perturbed rating matrix to the target domain.
However, these approaches are single-target, focusing on transferring knowledge from a source domain with rich data to improve recommendation quality in a target domain with sparse data. In some scenarios, both the source and target domains contain relatively richer information, such as ratings, review texts, user characteristics, and item attributes.
It is valuable to utilize the information from both domains to mutually improve recommendation performance.
In our method, we aim to simultaneously improve recommendation accuracy in both domains.

In addition, some methods  \cite{yu2021privacy,gao2019cross} only disclose either user or item representations to prevent attackers from inferring user-related data.
PPCDHWRec \cite{yu2021privacy} disentangles user characteristics in the source domain into domain-dependent and domain-independent features and then transfers these two types of features to the target domain.
During this process, only user features are disclosed, while item features remain hidden. Consequently, it's impossible to infer the original rating information.
Similarly, NATR \cite{gao2019cross} only shares item embeddings, which could protect user-related data.
However, the plaintext embedding is still at risk of being inferred \cite{chai2020secure}.
We add some noise to the disentangled embeddings before exchanging them across domains to relieve the risk of being inferred.

\section{Methodology}
In this part, we first present the definitions and notations used in this paper.
Subsequently, we provide a concise overview of the overall framework.
Following this, we introduce each module in detail.


\subsection{Definitions and Notations}
Suppose that we have two domains $A$ and $B$ with a shared user set $U=\{u_1,u_2,...,u_m\}$ (of size $m$) and 
different item sets $I^A = \{i^A_1, i^A_2,..., i^A_{n^A}\}$ (of size $n^A$), $I^B = \{i^B_1, i^B_2,..., i^B_{n^B}\}$ (of size $n^B$).
$A$ represents the source domain, and $B$ denotes the target domain.

For the input data, there are four modalities: user-item rating matrix, user review texts, item visual features, and item textual features.
Let $\textbf{R}^A\in \{0,1\}^{m\times n^A}$ and $\textbf{R}^B\in \{0,1\}^{m\times n^B}$ represent the binary user-item interaction matrices in domains $A$ and $B$, respectively.
We begin by aggregating interaction data within each domain to construct two heterogeneous graphs, denoted as $G^A = (U, I^A, S^A)$ and $G^B = (U, I^B, S^B)$.
These graphs serve as the foundation for learning user ID embeddings $\textbf{E}_u^A$, $\textbf{E}_u^B$, and item ID embeddings $\textbf{E}_i^A$, $\textbf{E}_i^B$ within the $A$ and $B$ domains, respectively.
Here, $S^A$ and $S^B$ represent the edge sets that capture the observed user-item interactions.
Let $\textbf{M}^A$ and $\textbf{M}^B$ denote user review embeddings, $\textbf{T}^A$ and $\textbf{T}^B$ indicate item textual embeddings, and $\textbf{V}^A$ and $\textbf{V}^B$ represent item visual embeddings in domains $A$ and $B$.

$\textbf{H}_u^A$ and $\textbf{H}_u^B$ represent user representations, while $\textbf{H}_i^A$ and $\textbf{H}_i^B$ indicate item representations in domains $A$ and $B$.
Given the user representations $\textbf{H}_u^A$ ($\textbf{H}_u^B$), we disentangle them into domain-specific embeddings $\textbf{P}_s^A$ ($\textbf{P}_s^B$) and domain-common embeddings $\textbf{P}_c^A$ ($\textbf{P}_c^B$).
At the same time, feature dropout is employed to generate augmented domain-specific embeddings $\tilde{\textbf{P}}_s^A$ ($\tilde{\textbf{P}}_s^B$) and domain-common embeddings $\tilde{\textbf{P}}_c^A$ ($\tilde{\textbf{P}}_c^B$).
Then, we utilize LDP to add noise to the decoupled embeddings and get the obfuscated embeddings $\textbf{Q}_s^A$ ($\textbf{Q}_s^B$) and $\textbf{Q}_c^A$ ($\textbf{Q}_c^B$), as well as the augmented obfuscated embeddings $\tilde{\textbf{Q}}_s^A$ ($\tilde{\textbf{Q}}_s^B$) and $\tilde{\textbf{Q}}_c^A$ ($\tilde{\textbf{Q}}_c^B$).
The goal of our method is to recommend Top-N items for all users in each domain.
The mathematical notations used in this paper are summarized
in Table \ref{notations}.

\begin{table}[h]
\centering
\caption{Notations.}\label{notations}
\begin{tabular}{c|c}
\hline
Symbols&Definitions and Notations\\
\hline
$*^A$, $*^B$&\makecell{domains A and B, \\e.g. $\textbf{R}^A$ represents the rating matrix in domain A} \\
U &user set\\
I & item set\\
m & the number of users\\
n & the number of items\\
$\textbf{R}$ & rating matrix\\
$G$ & heterogeneous graph\\
$\textbf{E}_u$ & user ID embeddings\\
$\textbf{E}_i$ & item ID embeddings\\
$\textbf{M}$ & user review embeddings\\
$\textbf{T}$ & item textual embeddings\\
$\textbf{V}$ & item visual embeddings\\
$\textbf{H}_u$ & user representations\\
$\textbf{H}_i$ & item representations\\
$\textbf{P}_s$ & domain-specific embeddings\\
$\tilde{\textbf{P}}_s$ & augmented domain-specific embeddings\\
$\textbf{P}_c$ & domain-common embeddings\\
$\tilde{\textbf{P}}_c$ & augmented domain-common embeddings\\
$\textbf{Q}_s$ & obfuscated domain-specific embeddings\\
$\tilde{\textbf{Q}}_s$ & augmented obfuscated domain-specific embeddings\\
$\textbf{Q}_c$ & obfuscated domain-common embeddings\\
$\tilde{\textbf{Q}}_c$ & augmented obfuscated domain-common embeddings\\
$\textbf{H}_u^*$ & comprehensive user preferences\\
\hline
\end{tabular}

\end{table}

\subsection{An overview of the proposed model}
\begin{figure*}[!htb] 
\centering 
\includegraphics[width=1.0\textwidth]{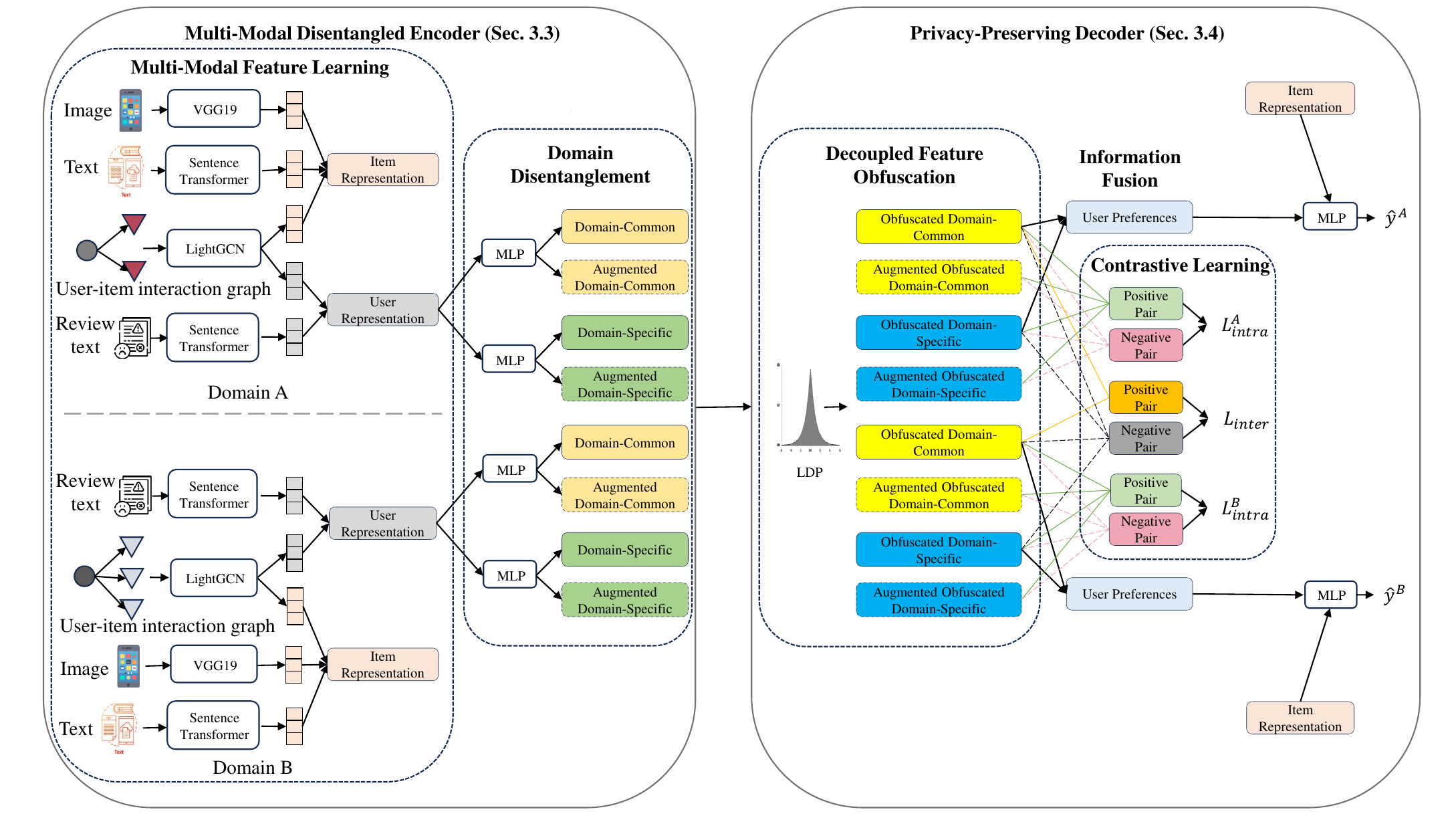}
\caption{The overall framework of P2M2-CDR.
It contains two modules: (1) Multi-Modal Disentangled Encoder, which first incorporates user-item rating matrix and multi-modal information w.r.t. review texts and visual and textual features to learn initial user 
and item representations and then disentangles user representations into domain-common and domain-specific embeddings. It contains multi-modal feature learning and domain disentanglement components. (2) Privacy-Preserving Decoder, which introduces local differential privacy (LDP) to safeguard user privacy. This module includes decoupled feature obfuscation, contrastive learning, and information fusion components.} 
\label{framework} 
\end{figure*}

We propose a privacy-preserving framework with multi-modal data for cross-domain recommendation (P2M2-CDR). 
Initially, the framework employs multi-modal data to decouple more informative domain-common and domain-specific embeddings.
Subsequently, it introduces local differential privacy (LDP) to prevent user privacy leakage.
Figure \ref{framework} provides an overview of the proposed framework.

This framework mainly includes two key modules:
\begin{itemize}
\item \textbf{Multi-Modal Disentangled Encoder}:
It contains two components:
(1) \textbf{Multi-Modal Feature Learning}: We incorporate multi-modal information, e.g., user-item interaction matrix, review texts, visual features, and text features, to learn user and item representations.
(2) \textbf{Domain Disentanglement}: We utilize MLP networks to disentangle user representations into more informative domain-common and domain-specific embeddings.
\item \textbf{Privacy-preserving Decoder}:
This module includes three components:
(1) \textbf{Decoupled Feature Obfuscation}: We introduce Laplace noise into disentangled embeddings to ensure the privacy protection of user data.
(2) \textbf{Contrastive Learning}: We introduce contrastive learning with domain-intra and domain-inter losses to regulate the separation and alignment of obfuscated decoupled embeddings.
(3) \textbf{Information Fusion}: We utilize different fusion methods to combine obfuscated domain-common and domain-specific embeddings into final user preferences.
\end{itemize}

\subsection{Multi-Modal Disentangled Encoder}
This module is designed to disentangle separate and informative domain-common and domain-specific embeddings.
\subsubsection{Multi-Modal Feature Learning}
In this section, we introduce multi-modal information, including a user-item interaction matrix, user review texts, and item visual and textual features, to learn initial user and item representations.

\textbf{ID Embeddings: } Motivated by the success of Graph Neural Networks (GNNs) that are good at modeling high-dimensional, complex relationships between users and items.
We introduce LightGCN \cite{he2020lightgcn}, a simple and lightweight graph model, to learn the user and item ID embeddings.
Firstly, we construct two heterogeneous graphs $G^A$ and $G^B$ to decipt the user-item interaction relationships in source domain $A$ and target domain $B$,
where nodes represent the user and item entities and edges show the relationship (whether the user and item interact) between entities.
In this paper, we use graph convolution and propagation layers in the LightGCN to encode the user and item ID embeddings according to the heterogeneous graphs $G^A$ and $G^B$.
We use $\textbf{E}_l^A$ (or $\textbf{E}_l^B$) to denote the ID embeddings at the $l$-th layer.
Specifically, the embeddings $\textbf{E}_0^A$ (or $\textbf{E}_0^B$) are randomly initialized.
Given the graph $G^A$, the $\textbf{E}_l^A$ could be calculated as follows,
\begin{equation}
\textbf{E}_l^A = (\textbf{D}^{-1/2}\textbf{A}\textbf{D}^{-1/2})\textbf{E}_{l-1}^A,
\end{equation}
where $\textbf{D}$ is a diagonal matrix and $\textbf{A}$ is an adjacency matrix.
After $l$ times of propagation, we can generate the final user ID embedding matrix $\textbf{E}_u^A$ and item ID embedding matrix $\textbf{E}_i^A$ by concatenating multiple embedding matrices from $\textbf{E}_0^A$ to $\textbf{E}_l^A$.
Similarly, we can obtain the final user ID embedding matrix $\textbf{E}_u^B$ and item ID embedding matrix $\textbf{E}_i^B$ in domain $B$.

\textbf{User and Item Multi-modal Embeddings:}
For every user, we aggregate all review texts associated with the items they have rated to create the user-specific review text.
Then we use the pre-trained model Sentence Transformer \cite{reimers2019sentence} and a MLP layer to generate user primitive review embeddings $\textbf{M}^A$ (or $\textbf{M}^B$). 
Similarly, we concatenate the item's title, categories, and description as item textual features.
These item textual features are then processed through the Sentence Transformer model and a MLP layer to obtain the final item textual embeddings $\textbf{T}^A$ (or $\textbf{T}^B$).
For item visual features, we utilize VGG19 \cite{simonyan2014very} in conjunction with a MLP layer to learn the item visual embeddings $\textbf{V}^A$ (or $\textbf{V}^B$).
  
\textbf{User and Item Representations:}
Finally, we concatenate the user ID embeddings $\textbf{E}_u^A$ and user review embeddings $\textbf{M}^A$ to obtain user representations $\textbf{H}_u^A$ in domain $A$,
\begin{equation}
\textbf{H}_u^A = f(\textbf{E}_u^A, \textbf{M}^A),
\end{equation}
where $f$ denotes the concatenation function.
In a similar manner, we can get user representations $\textbf{H}_u^B$ in domain $B$.

For item representations, we concatenate item ID embeddings $\textbf{E}_i^A$, item visual embeddings $\textbf{V}^A$, and item textual embeddings $\textbf{T}^A$ to learn item representations $\textbf{H}_i^A$,
 \begin{equation}
\textbf{H}_i^A = f(\textbf{E}_i^A, \textbf{V}^A, \textbf{T}^A).
\end{equation}

Similarly, we can get item representations $\textbf{H}_i^B$ in domain $B$.
\subsubsection{Domain Disentanglement}
Here, we decouple user representations $\textbf{H}_u^A$  into more informative domain-specific embeddings $\textbf{P}_s^A$  and domain-common embeddings $\textbf{P}_c^A$ via two MLP layers,  
 \begin{equation}
\textbf{P}_s^A = MLP(\textbf{H}_u^A; \Theta_s^A);   \qquad  \textbf{P}_c^A = MLP(\textbf{H}_u^A; \Theta_c^A).
\end{equation}
Simultaneously, it disentangles $\textbf{H}_u^B$ into $\textbf{P}_s^B$ and $\textbf{P}_c^B$,
 \begin{equation}
\textbf{P}_s^B = MLP(\textbf{H}_u^B; \Theta_s^B);   \qquad  \textbf{P}_c^B = MLP(\textbf{H}_u^B; \Theta_c^B),
\end{equation}
where $\Theta_s^A$, $\Theta_c^A$, $\Theta_s^B$ and $\Theta_c^B$ are all the parameters for the respective MLP layers.

To enhance the informativeness of disentangled representations, we employ the feature dropout strategy introduced by Zhou \cite{zhou2020s3} to generate augmented decoupled embeddings in both domains, denoted as $\tilde{\textbf{P}}_s^A$, $\tilde{\textbf{P}}_c^A$ in domain A, and $\tilde{\textbf{P}}_s^B$, $\tilde{\textbf{P}}_c^B$ in domain B.

\subsection{Privacy-preserving Decoder}
In this section, we encounter three key challenges:
(1) How to obfuscate domain-common and domain-specific embeddings to protect user-related information?
(2) How to ensure that obfuscated disentangled embeddings don't contain redundant information within their domains and include shared features across domains to enhance informativeness?
(3) How to aggregate obfuscated domain-common and domain-specific embeddings to learn comprehensive and diversified user preferences?
To address these issues, we introduce a decoupled feature obfuscation component, a contrastive learning component, and an information fusion component.

\subsubsection{Decoupled Feature Obfuscation}
In this part, disentangled embeddings from both the source and target domains are shared when introducing contrastive learning to achieve alignment and separation. 
However, doing so directly carries the potential risk of inferring users' actual data \cite{chai2020secure}.
To mitigate this risk and prevent potential leakage of user-sensitive data, we employ local differential privacy (LDP) techniques to obscure the true distribution of these disentangled embeddings.
More specifically, we add Laplace noise to these disentangled embeddings,
which not only safeguards against external attackers intercepting the original decoupled embeddings but also prevents the other domain from inferring the user's sensitive information from these embeddings.
The obfuscated disentangled embeddings could be obtained as follows,

\begin{equation}
\begin{split}
\textbf{Q}_s^A = \textbf{P}_s^A + La(\mu, \lambda);   \qquad  \textbf{Q}_c^A = \textbf{P}_c^A + La(\mu, \lambda);  \\
\tilde{\textbf{Q}_s^A} = \tilde{\textbf{P}_s^A} + La(\mu, \lambda);   \qquad  \tilde{\textbf{Q}_c^A} = \tilde{\textbf{P}_c^A} + La(\mu, \lambda), 
\end{split}
\end{equation}

where $\mu$ and $\lambda$ represent the mean value and standard deviation of Laplace noise. Here, we set $\mu=0$.

Similarly, we can get the disturbing domain-specific embeddings $\textbf{Q}_s^B$, $\tilde{\textbf{Q}_s^B}$ and domain-common embeddings $\textbf{Q}_c^B$, $\tilde{\textbf{Q}_c^B}$ in domain $B$.

\subsubsection{Contrastive Learning}
In the last subsection, we obtain obfuscated domain-common and domain-specific features.
However, we couldn't guarantee that these embeddings include different aspects within each domain.
Moreover, we cannot ensure whether the obfuscated domain-common features in the source and target domains are indeed similar.
Here, we introduce contrastive learning-based domain-intra and domain-inter losses to solve the above challenges.
The core idea is to bring positive sample pairs closer together and push negative sample pairs farther apart.

To promote the separation of obfuscated domain-common and domain-specific embeddings, 
we treat variants of the same disentangled embedding as positive pairs and variants of different disentangled embeddings as negative pairs.
 For example, $(\textbf{Q}_c^A, \tilde{\textbf{Q}}_c^A)$ is a positive sample pair, $(\textbf{Q}_c^A, \textbf{Q}_s^A)$ and $(\textbf{Q}_c^A,  \tilde{\textbf{Q}}_s^A)$ are negative sample pairs.
The domain-intra loss is defined as follows,
 
  \begin{equation}\small
  \begin{aligned}
l_{intra}^p &= exp(f(\textbf{Q}_c^A, \tilde{\textbf{Q}}_c^A)/\tau)+exp(f(\textbf{Q}_s^A, \tilde{\textbf{Q}}_s^A)/\tau);\\
l_{intra}^n &= exp(f(\textbf{Q}_c^A, \textbf{Q}_s^A)/\tau)+exp(f(\textbf{Q}_c^A, \tilde{\textbf{Q}}_s^A)/\tau)\\
     &+exp(f(\textbf{Q}_s^A, \tilde{\textbf{Q}}_c^A)/\tau)+exp(f( \tilde{\textbf{Q}}_s^A, \tilde{\textbf{Q}}_c^A)/\tau);\\
L_{intra}^A &= -log\frac{l_{intra}^p}{l_{intra}^p+l_{intra}^n},
 \end{aligned}
 \end{equation}
 
where $f$ represents the similarity function and $\tau$ is a trainable hyper-parameter of temperature.
Similarly, we could get the domain-intra loss $L_{intra}^B$ in domain $D^B$.
 
For the domain-inter loss, our focus is on aligning $\textbf{Q}_c^A$ and $\textbf{Q}_c^B$ to ensure they contain domain-common information while simultaneously separating $\textbf{Q}_s^A$ and $\textbf{Q}_s^B$ to include domain-specific information.
 We consider $(\textbf{Q}_c^A, \textbf{Q}_c^B)$ as a positive sample pair,  $(\textbf{Q}_c^A, \textbf{Q}_s^B)$, $(\textbf{Q}_c^B, \textbf{Q}_s^A)$  and $(\textbf{Q}_s^A, \textbf{Q}_s^B)$ as negative sample pairs.
 The domain-inter loss could be calculated as follows, 
 {\footnotesize
 \begin{equation}
 \begin{aligned}
l_{inter}^p &= exp(f(\textbf{Q}_c^A, \textbf{Q}_c^B)/\tau)\\
l_{inter}^n &= exp(f(\textbf{Q}_c^A, \textbf{Q}_s^B)/\tau)+exp(f(\textbf{Q}_c^B, \textbf{Q}_s^A)/\tau)\\
&+exp(f(\textbf{Q}_s^A, \textbf{Q}_s^B)/\tau)\\
 L_{inter} &= -log\frac{l_{inter}^p}{l_{inter}^p+l_{inter}^n}.
 \end{aligned}
 \end{equation}
 }
 Finally, the objective of contrastive learning is as follows,
 \begin{equation}
 L_C =  L_{intra}^A+ L_{intra}^B+ L_{inter}.
 \end{equation} 
 
 \subsubsection{Information Fusion}
Domain-shared and domain-specific embeddings constitute two fundamental components of user preferences, and it is imperative to integrate them in a rational and effective manner to comprehensively capture user preferences.
Therefore, we employ three fusion methods, namely element-wise sum, concatenation, and attention mechanisms, to aggregate individual embeddings into comprehensive user preferences.
Following experimental validation in Section IV, we ultimately selected the element-wise sum method to generate comprehensive user preferences,
\begin{equation}
H_u^{A*} = g(\textbf{Q}_c^A,\textbf{Q}_s^A),
\end{equation}
where g denotes the element-wise sum function.
Similarly, we could get the comprehensive user preferences $H_u^{B*}$ in domain B.
 
\subsection{Model Training and Optimization}
In this section, we feed both the user preferences $H_u^{A*}$ ($H_u^{B*}$) and item representations $H_i^A$ ($H_i^B$) into an MLP to predict the probability of a user clicking on a certain item.
We aim to minimize the following loss function,
\begin{equation}
L_{prd}^A = \sum_{r\in r^{A+}\cup r^{A-}}l(\hat{r},r),
\end{equation}
where $l$ denotes the cross-entropy loss function.
The total loss function is defined as follows,
\begin{equation}
L = L_{prd}^A + L_{prd}^B +\alpha L_C,
\label{loss function}
\end{equation} 
where $\alpha$ is the weight parameter for the contrastive learning loss.

\subsection{Privacy Analysis}
In this section, we mainly pay attention to the privacy analysis of our framework P2M2-CDR. 
Our work aims to prevent the leakage of user privacy information, which is a significant concern in existing CDR methods.

In our approach, data is stored within the central servers of each company (domain). These two domains communicate with each other as we introduce contrastive learning-based domain-intra and domain-inter losses to regulate the alignment and separation of disentangled embeddings.

Specifically, during the model's training, we encourage the common features of these two domains to become increasingly similar while enhancing the distinctiveness of specific features.
Throughout this process, these two domains exchange disentangled embeddings, which carries the risk of inferring users' sensitive data. 
For example, attackers may attempt to infer users' cross-domain behaviors or preferences from information shared across domains.
Consequently, we employ the local differential privacy (LDP) method to conceal these disentangled embeddings. This is achieved by introducing noise that adheres to the Laplace distribution.

Within LDP, a higher standard deviation ($\lambda$), which controls the noise strength, can more effectively safeguard sensitive data, thereby reducing the risk of data leakage. However, it may lead to a reduction in recommendation performance. Hence, it is essential to set an appropriate value for $\lambda$ to strike a balance between recommendation accuracy and privacy protection.
\section{Experiments}
In this section, in order to evaluate the performance of our proposed model, P2M2-CDR, we conduct a series of comprehensive experiments on four publicly available subsets from the Amazon dataset to answer the following questions.
\begin{itemize}
\item RQ1: Does our model achieve superior performance compared to other state-of-the-art baseline methods?
\item RQ2: How do different components, such as contrastive learning, decoupled feature obfuscation, domain-specific information, domain-common information, and multi-modal information influence the outcomes of our model?
\item RQ3: Are the embeddings we've acquired genuinely disentangled?
\item RQ4: How does our model's performance vary with different hyper-parameters?
\item RQ5: How does the performance of our model change with varying information fusion methods?
 \end{itemize}
 
\subsection{Experimental Settings}
\subsubsection{Datasets}
We conduct comprehensive experiments on four real-world benchmark subsets from Amazon dataset\footnote{https://cseweb.ucsd.edu/~jmcauley/datasets/amazon/links.html}: Cell Phones and Accessories (\textbf{Phone}), Electronics (\textbf{Elec}), Sports and Outdoors (\textbf{Sport}), and Clothing, Shoes and Jewelry (\textbf{Cloth}).
We combine them into four CDR scenarios: \textbf{Phone\&Elec}, \textbf{Phone\&Sport}, \textbf{Sport\&Cloth}, and \textbf{Elec\&Cloth} and extract common users for each pair of datasets.
For these four datasets, we transform the explicit ratings into implicit feedback. Specifically, we discretize the ratings into binary values of 0 and 1 to 
indicate whether the user has interacted with the item or not. 
For each observed user-item interaction, we randomly select an item that the user has not interacted with before as a negative sample.
To ensure data quality and alleviate sparsity issues, we apply filtering criteria to remove records with less than 5 interactions for all users and items across both domains.
Table \ref{tab2} provides an overview of the basic statistics of these preprocessed datasets.

\begin{table}[h]
\centering
\caption{Statistic of the datasets for four CDR scenarios.}\label{tab2}
\begin{tabular}{l l l l l l}
\hline
Datasets & Users & Items & Training & Test & Density\\
\hline
Phone & 22998 &38800 & 163267 & 22998 & 0.0208\%\\
Elec & 22998 & 77187 & 341031 & 22998 & 0.0205\%\\
\hline
Phone & 5902 & 18635 & 50210 & 5902 & 0.051\%\\
Sport & 5902 & 29180 & 61374 & 5902 & 0.039\%\\
\hline
Sport & 12965 & 46868 & 121711 & 12965 & 0.022\%\\
Cloth & 12965 & 62343 & 121502 & 12965 & 0.017\%\\
\hline
Elec & 19754 & 69362 & 259528 & 19754 & 0.020\%\\
Cloth & 19754 & 77003 & 159981 & 19754 & 0.012\%\\
\hline
\end{tabular}

\end{table}

\subsubsection{Evaluation Metrics}
Motivated by BiTGCF \cite{liu2020cross} and DIDA-CDR \cite{zhu2023domain}, we also use the leave-one-out method to evaluate the model's performance.
For each user, we randomly select one sample to create the test set, with the remaining samples forming the training set.
Following the NeuMF \cite{he2017neural}, for each test user, we randomly select 99 items that the user has not interacted with as negative samples and the ground-truth user-item interaction as the positive sample. Then, we employ the P2M2-CDR model to predict scores for these 100 candidate items to perform the ranking. The evaluation of recommendation performance relies on two metrics: Hit Ratio (HR) and Normalized Discounted Cumulative Gain (NDCG).
\subsubsection{Baseline Methods}
To verify the effectiveness of our model, we compare the performance of P2M2-CDR with three sets of representative baseline methods: Single-Domain Recommendation, 
Cross-Domain Recommendation and Privacy-Preserving Cross-Domain Recommendation.

\noindent \textbf{Single-Domain Recommendation}
\begin{itemize}
\item \textbf{NeuMF} \cite{he2017neural} combines collaborative filtering and neural network techniques to capture user-item interactions and make accurate predictions.
\item \textbf{LightGCN} \cite{he2020lightgcn} is a simple GCN model that directly propagates user and item embeddings through the user-item interaction graph without introducing complex operations or auxiliary information.
 \end{itemize}
 
 \noindent \textbf{Cross-Domain Recommendation}
\begin{itemize}
\item \textbf{PTUPCDR} \cite{zhu2022personalized} proposes a framework focusing on personalized user preference transfer via a meta-network.
\item \textbf{DDTCDR} \cite{li2020ddtcdr} proposes a method that aims to learn an orthogonal mapping function to transfer user preferences across domains and provide recommendations for both domains. 
In addition to user-item rating information, it also utilizes user and item features to improve recommendation accuracy.
\item \textbf{DCCDR} \cite{zhang2023disentangled} proposes a framework aiming at disentangling domain-invariant and domain-specific representations and then using GNN to learn high-order relationships to enrich these representations.
\item \textbf{DisenCDR} \cite{cao2022disencdr} focuses on disentangling user preferences into domain-specific and domain-shared information and transferring the domain-shared knowledge across domains.
\item \textbf{BiTGCF} \cite{liu2020cross} proposes a method that conducts bidirectional high-order information transfer to enhance the performance of graph collaborative filtering-based cross-domain recommendation.
 \end{itemize}
 
 \noindent \textbf{Privacy-Preserving Cross-Domain Recommendation}
 \begin{itemize}
\item \textbf{PriCDR} \cite{chen2022differential} proposes a privacy-preserving CDR framework that utilizes Differential Privacy (DP) technology to publish the rating matrix in the source domain and then transfers the published matrix to the target domain.
\item \textbf{P2FCDR} \cite{chen2023win} proposes a privacy-preserving federated framework that learns an orthogonal mapping matrix to transform embeddings across domains and apply local differential privacy technique on the transformed embeddings to protect user privacy.
  \end{itemize}
  
 \subsubsection{Parameter Settings}
We implement the P2M2-CDR model using Python with Pytorch framework, all baseline methods are conducted based on their GitHub source code and carefully adjusted the hyperparameters.
The optimal hyperparameters are obtained by optimizing the loss function \eqref{loss function} using the Adam optimizer with a learning rate of 0.001.
For the disentangled encoder, we use a two-layer fully connected network with dimensions 128 and 64, respectively, and we could obtain the disentangled embeddings with dimensions 64.
Considering the trade-off between recommendation performance and privacy protection, we set $\lambda$ to 0.01. Simultaneously, the weight of the contrastive learning loss is set to 0.001.
We set the batch size to 512 and the number of epochs to 200. To prevent overfitting, we apply batch normalization, dropout, and early stopping techniques.
 
 \subsection{Performance Evaluation (RQ1)}
We evaluate the performance of P2M2-CDR and the baselines using commonly used evaluation metrics w.r.t. HR and NDCG and we set the length of recommendation lists as 10. The results are shown in Table \ref{comparison} and Table \ref{comparison1}.
Since PTUPCDR and PriCDR are designed for single-target cross-domain recommendation, we only present the results in the target domain.

 \begin{table*}[!htb]
\centering
\caption{Overall comparison on scenarios: Phone\&Elec and Phone\&Sport. The optimal performance is highlighted using bold fonts, and the second-best performance
is denoted by underlines.}\label{comparison}
\begin{tabular}{c c c c c c c c c}
\hline
\multirow{2}*{} &\multicolumn{4}{c}{Phone / Elec}&\multicolumn{4}{c}{Phone / Sport}\\
\cmidrule(r){2-5}\cmidrule(r){6-9}
 &HR@10&NDCG@10&HR@10&NDCG@10&HR@10&NDCG@10&HR@10&NDCG@10\\
\hline
\multicolumn{9}{l}{\textbf{Single domain recommendations} }\\
 NeuMF & 0.5093 &0.3357   & 0.4412 &0.2848  &0.3531 &0.2296  & 0.3053 & 0.1808 \\
 LightGCN  & 0.5124 & 0.3478   & 0.4498 & 0.2901  & 0.3624 & 0.2307  & 0.3214 & 0.1956 \\
\hline
\multicolumn{9}{l}{\textbf{Cross domain recommendations} }\\
 PTUPCDR &--&  --  & 0.4813 & 0.2171  & -- &  -- & 0.4873 & 0.2228 \\
 BiTGCF  &0.5360 &0.3568   &0.4971 &0.1167  &0.5076 & 0.2740  & 0.5290 &0.3042 \\
DisenCDR  & 0.5423 & 0.3751   & 0.5903 & 0.3935  & 0.5314 & 0.3476  & 0.5323 & 0.3315 \\
DDTCDR  & 0.7063 &0.4228   &0.6539 &0.4023  & 0.5504 &0.3601  &0.5672 &0.3521 \\
DCCDR  & \underline{0.7234} & 0.4389   & \underline{0.6612} & \underline{0.4110}  & \underline{0.6606} & \underline{0.3887}  & \underline{0.6325} & \underline{0.3894} \\
\hline
\multicolumn{9}{l}{\textbf{Privacy-Preservig Cross-domain recommendations} }\\
PriCDR &-- &-- &0.5515 &0.3688 &-- &-- & 0.5315&0.3789\\
P2FCDR & 0.7156 & \underline{0.4414} & 0.6478 & 0.3967 & 0.6302 & 0.3838 & 0.6270 & 0.3801\\

\hline
\multicolumn{9}{l}{\textbf{Ours} }\\
 P2M2-CDR&\textbf{0.8570} & \textbf{0.5335}   & \textbf{0.7803} & \textbf{0.4890}  & \textbf{0.9281} & \textbf{0.5827}  & \textbf{0.9151}&\textbf{0.5998}\\
 Imp&13.36\%&9.21\%&11.91\%&7.8\%&26.75\%&19.40\%&28.26\%&21.04\%\\
\hline 
\end{tabular}
\end{table*}

 \begin{table*}[!htb]
\centering
\caption{Overall comparison on scenarios: Sport\&Cloth and Elec\&Cloth. The optimal performance is highlighted using bold fonts, and the second-best performance
is denoted by underlines.}\label{comparison1}
\begin{tabular}{c c c c c c c c c}
\hline
\multirow{2}*{} &\multicolumn{4}{c}{Sport / Cloth}&\multicolumn{4}{c}{Elec / Cloth}\\
\cmidrule(r){2-5}\cmidrule(r){6-9}
&HR@10&NDCG@10&HR@10&NDCG@10&HR@10&NDCG@10&HR@10&NDCG@10\\
\hline
\multicolumn{9}{l}{\textbf{Single domain recommendations} }\\
 NeuMF & 0.3495 & 0.2131   &0.2448 &0.1523  & 0.4007 & 0.2562  & 0.2435 &0.1315 \\
LightGCN  & 0.3562 & 0.2257   & 0.2592 & 0.1645  & 0.4120 & 0.2635  & 0.2612 & 0.1478 \\
\hline
\multicolumn{9}{l}{\textbf{Cross domain recommendations} }\\
PTUPCDR &--  &  --  & 0.4821 & 0.2191  & -- & --  & 0.4828 & 0.2188 \\
BiTGCF  & 0.5409 &0.3069   &0.5491 & 0.2943  &0.6366 &0.3429  & 0.5975 &0.3479 \\
DisenCDR  & 0.5623 & 0.3351   & 0.5812 & 0.3301  & 0.6414 & 0.3517  & 0.6070 & 0.3467 \\
DDTCDR  &0.5912 &0.3652   &0.6078 &0.3554  & 0.6514 & 0.3552  &0.6198 & 0.3592 \\
DCCDR  & \underline{0.7342} & \underline{0.4424}   & \underline{0.6946} & \underline{0.4023}  & \underline{0.7069} & \underline{0.4112} & \underline{0.6578} & \underline{0.3612} \\
\hline
\multicolumn{9}{l}{\textbf{Privacy-Preservig Cross-domain recommendations} }\\
PriCDR &-- &-- &0.5330 &0.3237 & --&-- &0.5749 &0.3412\\
P2FCDR & 0.7196 & 0.4276 & 0.6903 & 0.3778 & 0.6845 & 0.3672 & 0.6503 & 0.3579\\
\hline
\multicolumn{9}{l}{\textbf{Ours}}\\
P2M2-CDR&\textbf{0.8964} & \textbf{0.5491}   & \textbf{0.8642} &\textbf{0.5236} &\textbf{0.7893} & \textbf{0.4872}  & \textbf{0.8479} &\textbf{0.5057}\\
Imp&16.22\%&10.67\%&16.96\%&12.13\%&8.24\%&7.6\%&19.01\%&14.55\%\\
\hline 
\end{tabular}
\end{table*}

\begin{itemize}
\item Our model, P2M2-CDR, outperforms other baselines, achieving the best performance with up to a 28.26\% improvement over the best baseline, DCCDR, on HR@10, and a 21.04\% improvement on NDCG@10.
Additionally, despite our method introducing some noise to protect user privacy, it still surpasses other cross-domain recommendation methods. These results indicate that the proposed privacy-preserving framework could enhance recommendation performance in both the source and target domains simultaneously while safeguarding user privacy.
\item Cross-domain recommendation methods outperform single-domain recommendation models, especially in domains with sparse data.
For instance, DisenCDR demonstrates superior performance compared to LightGCN, achieving an average improvement of 33.39\% and 18.23\% in HR@10 and NDCG@10, respectively, on the Cloth dataset.
This demonstrates that cross-domain recommendation methods are effective in addressing the data-sparsity problem.
\item GNN-based methods outperform non-graph methods, such as LightGCN vs NeuMF, and DCCDR vs DisenCDR. This demonstrates that incorporating high-order neighbor information can improve model accuracy.
\item Integrating user and item text features could improve model performance, such as DDTCDT vs BiTGCF. 
In addition, multi-modal features play an important role in improving recommendation accuracy, such as our model P2M2-CDR vs DDTCDR.
\item Disentanglement-based dual-target CDR methods surpass other dual-target CDR methods, such as DisenCDR vs BiTGCF, and P2M2-CDR vs P2FCDR.  
This is because disentanglement-based methods typically decouple user preferences into domain-common and domain-specific features and only transfer the common features, which could avoid negative transfer.

 \end{itemize}
 
 \subsection{Ablation Studies (RQ2)}
To assess the effectiveness of each component in P2M2-CDR, we conducted ablation experiments on four commonly used subsets from the Amazon dataset.
We created eight variants of P2M2-CDR by removing specific components.
\begin{itemize}
\item w/o rev: It removes user review texts when learning comprehensive user representations.
\item w/o vis: It removes item visual features when learning comprehensive item representations.
\item w/o txt: It removes item textual features when learning comprehensive item representations.
\item w/o com: It removes obfuscated domain-common features for recommendation.
\item w/o spe: It removes obfuscated domain-specific features for recommendation.
\item w/o intra: It eliminates the domain-intra loss.
\item w/o inter: It removes the domain-inter loss.
\item w/o obf: It deletes the decoupled feature obfuscation module.
\end{itemize}
The results of the ablation studies are presented in Table \ref{ablation} and Table \ref{ablation1}.
\begin{table*}[htb]
\centering
\caption{Ablation studies on scenarios: Phone\&Elec and Phone\&Sport.}\label{ablation}
\begin{tabular}{c c c c c c c c c}
\hline
\multirow{2}*{Variants} &\multicolumn{4}{c}{Phone / Elec}&\multicolumn{4}{c}{Phone / Sport}\\
\cmidrule(r){2-5}\cmidrule(r){6-9}
 &HR@10&NDCG@10&HR@10&NDCG@10&HR@10&NDCG@10&HR@10&NDCG@10\\
\hline
w/o rev&0.5688&0.3627&0.5620&0.3685&0.5430&0.3200&0.5408&0.2881\\
w/o vis&0.7180&0.4656&0.6648&0.4323 &0.5393&0.3286&0.5713&0.3304\\
w/o txt&0.6563&0.4209&0.6161&0.3951&0.8035&0.5037&0.8195&0.4867\\
w/o com&0.8307 &0.5156 &0.7512 & 0.4738&0.8207 & 0.4834&0.8233 &0.4856 \\
w/o spe&0.8329&0.5145&0.7787&0.4869&0.8677&0.5181&0.8926&0.5578\\
w/o intra&0.8342 &0.5046 & 0.7489& 0.4458&0.9071&0.5386&0.9002&0.5486\\
w/o inter&0.8352 &0.5174 &0.7448 &0.4449 &0.9158&0.5771&0.9068&0.5950\\
w/o obf& 0.8526 &\textbf{0.5356}  & \textbf{0.7921} &\textbf{0.4909}  & \textbf{0.9397}&\textbf{0.6154}&\textbf{0.9437}&\textbf{0.6368}\\			
P2M2-CDR&\textbf{0.8570} & 0.5335   & 0.7803 & 0.4890  & 0.9281 & 0.5827  & 0.9151&0.5998\\
\hline
\end{tabular}
 \end{table*}

\begin{table*}[htb]
\centering
\caption{Ablation studies on scenarios: Sport\&Cloth and Elec\&Cloth.}\label{ablation1}
\begin{tabular}{c c c c c c c c c}
\hline
\multirow{2}*{Variants} &\multicolumn{4}{c}{Sport / Cloth}&\multicolumn{4}{c}{Elec / Cloth}\\
\cmidrule(r){2-5}\cmidrule(r){6-9}
 &HR@10&NDCG@10&HR@10&NDCG@10&HR@10&NDCG@10&HR@10&NDCG@10\\
\hline
w/o rev&0.5019&0.2842&0.4969&0.2801&0.5445&0.3456&0.4141&0.2436\\
w/o vis&0.6495&0.3859&0.5650&0.3304&0.6018&0.3860&0.5561&0.3223\\
w/o txt&0.7874&0.4591&0.7498&0.4338&0.6238&0.3882&0.6033&0.3459\\
w/o com& 0.8801&0.5342 &0.8344 & 0.4894& 0.7689&0.4681 &0.7811 &0.4464 \\
w/o spe&0.8684&0.5178&0.8322&0.4913&0.7834&0.4697&0.8312&0.4723\\
w/o intra&0.8356 &0.5034 &0.8321 &0.4980 &0.7635 &0.4589 &0.8026 &0.4620 \\
w/o inter&0.8440 &0.5059 &0.7962 &0.4657 &0.7759&0.4783&0.7728&0.4412\\
w/o obf&\textbf{0.8972}&\textbf{0.5495}&\textbf{0.8728}&\textbf{0.5268}& \textbf{0.8093}&\textbf{0.4973}&\textbf{0.8687}&\textbf{0.5148} \\
P2M2-CDR&0.8964 & 0.5491  & 0.8642 &0.5236  & 0.7893 & 0.4872  & 0.8479 & 0.5057\\
\hline
\end{tabular}
\end{table*}
Based on the above results, we observe that each component of the overall model plays a crucial role. 
\begin{itemize}
    \item Without user review texts, the performance of `w/o rev' declines on average by 34\% and 22\% in terms of HR@10 and NDCG@10 compared to P2M2-CDR. This highlights the significance of user review texts in disentangling domain-common and domain-specific features and modeling comprehensive user representations.
    \item The model `w/o vis', which eliminates the item visual features, experiences a drop in performance by an average of 25.21\% in terms of HR@10 and 16.11\%  in terms of NDCG@10 on four cross-domain recommendation scenarios. Similarly, the model `w/o txt', which eliminates the item textual features, experiences an average drop of 15.23\% and 10.46\% in terms of HR@10 and NDCG@10.
    These results demonstrate the significance of incorporating both visual and textual features of items in the process of learning comprehensive item representations.
    \item Comparing P2M2-CDR with `w/o com' and `w/o spe', we could see that obfuscated disentangled domain-common and domain-specific features play a vital role in predicting user preferences.
    \item The inferior performance of models `w/o intra' and `w/o inter' further shows the significant contributions of both domain-intra and domain-inter losses to the final outcome.
    \item It is worth noting that the model `w/o obf', without the addition of noise, outperforms P2M2-CDR. However, this approach may pose a privacy threat by potentially exposing user data. We should carefully balance the trade-off between ensuring privacy protection and achieving optimal model performance.
\end{itemize}

\subsection{Visualization of Obfuscated Disentangled Embeddings (RQ3)}
\begin{figure}[!t]
\centering
\subfloat[]{\includegraphics[width=1.6in]{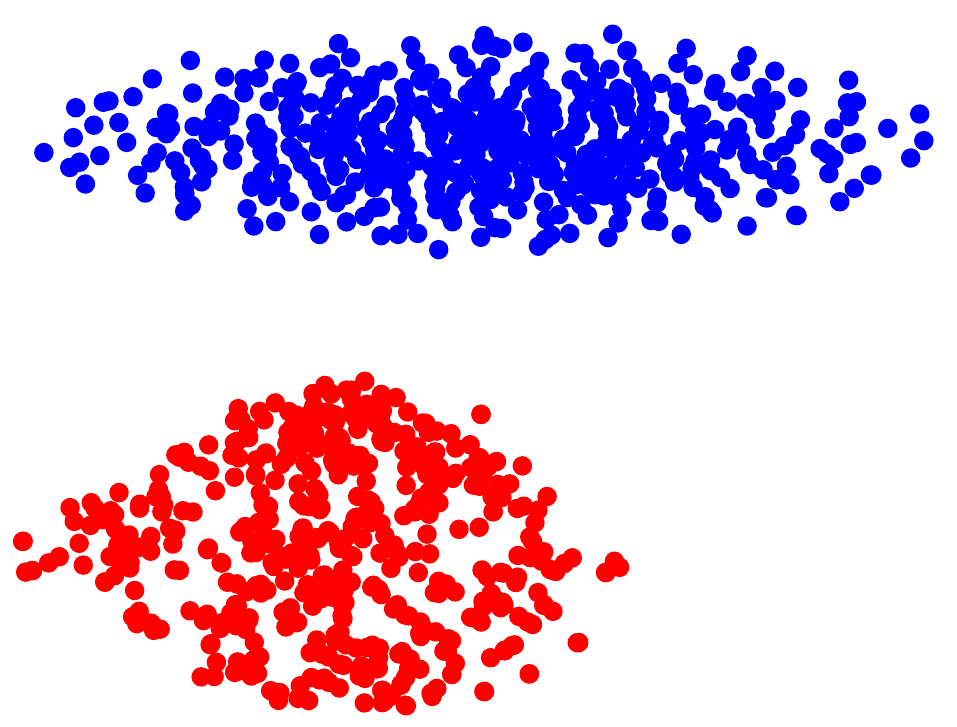}%
\label{disen_c_s}}
\hfil
\subfloat[]{\includegraphics[width=1.6in]{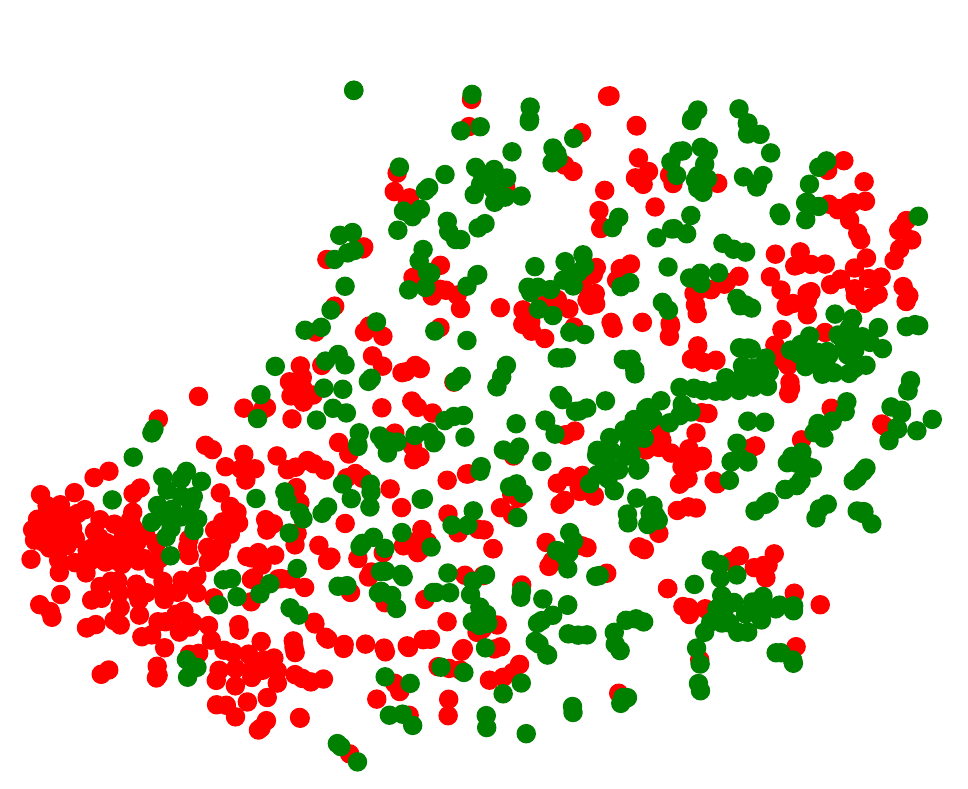}%
\label{disen_source_target_c}}
\caption{ Visualization of user obfuscated disentangled embeddings in the scenario: \textbf{Phone\&Sport}. (a) Red points represent obfuscated domain-common embeddings and blue points indicate obfuscated domain-specific embeddings in the source domain (\textbf{Phone}); 
(b) Red points represent obfuscated domain-common embeddings in the source domain (\textbf{Phone}) and green points show the obfuscated domain-common embeddings in the target domain (\textbf{Sport}).}
\label{visualize}
\end{figure}

\begin{figure}[!t]
\centering
\subfloat[]{\includegraphics[width=1.6in]{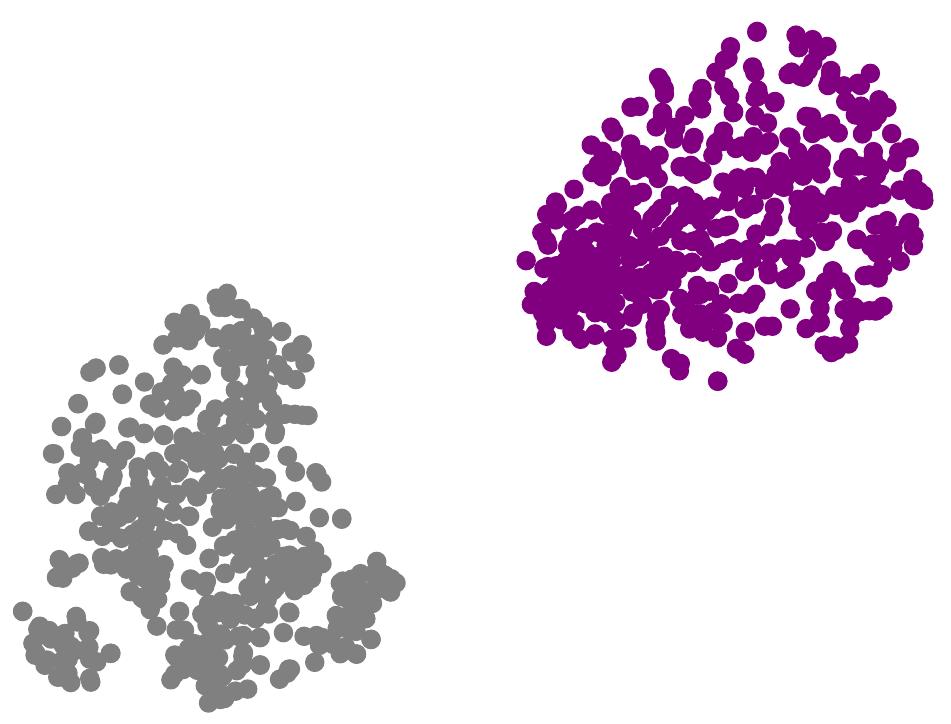}%
\label{sport_cloth_disen_c_s}}
\hfil
\subfloat[]{\includegraphics[width=1.6in]{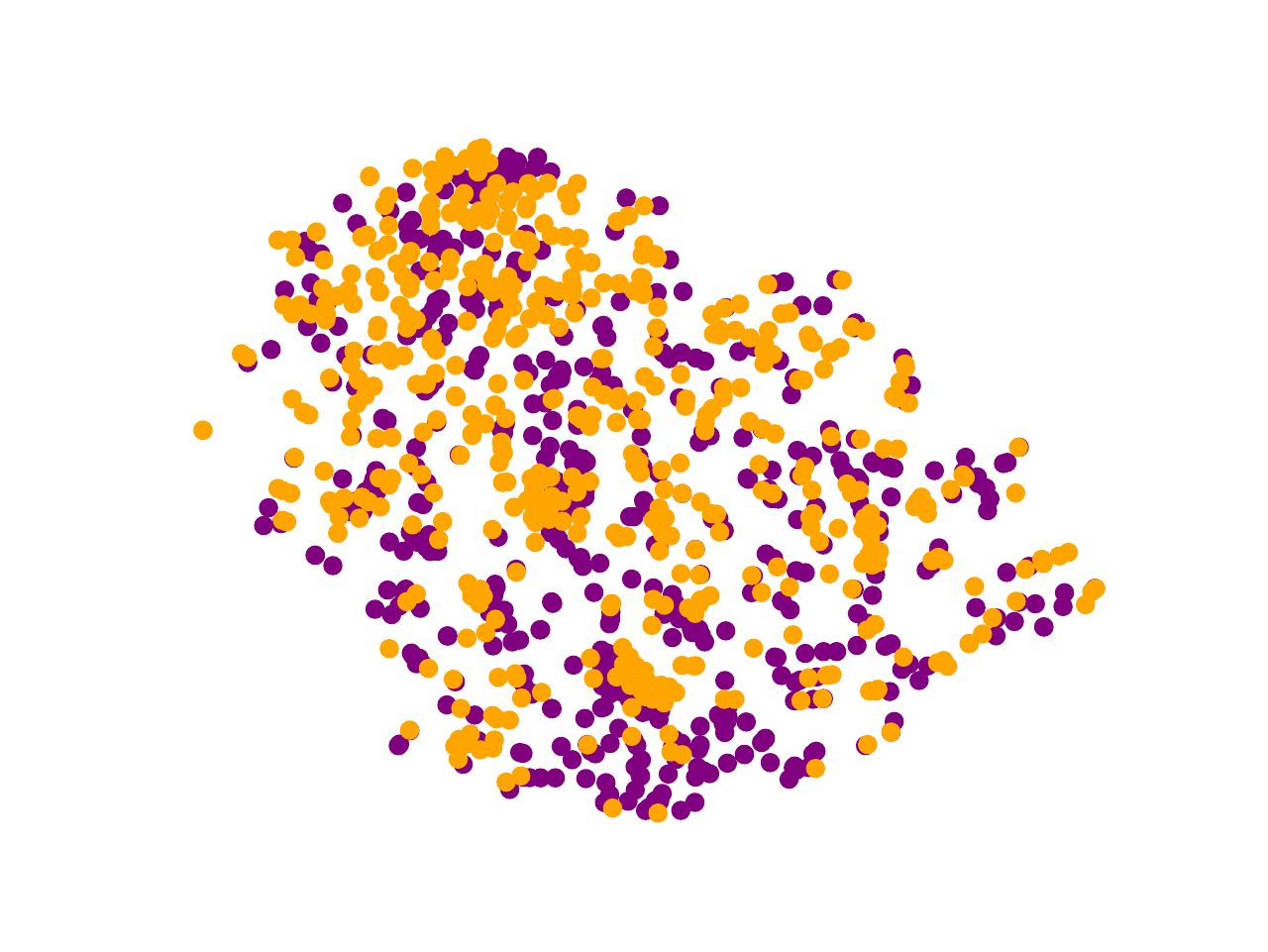}%
\label{sport_cloth_disen_source_target_c}}
\caption{ Visualization of user obfuscated disentangled embeddings in the scenario: \textbf{Sport\&Cloth}. (a) Purple points represent obfuscated domain-common embeddings and gray points indicate obfuscated domain-specific embeddings in the source domain (\textbf{Sport}); 
(b) Purple points represent obfuscated domain-common embeddings in the source domain (\textbf{Sport}) and orange points show the obfuscated domain-common embeddings in the target domain (\textbf{Cloth}).}
\label{sport_cloth_visualize}
\end{figure}

 To determine whether obfuscated domain-common and domain-specific features are distinct and contain diverse information, we visualize these embeddings using t-SNE \cite{van2008visualizing} 
which is a data visualization technique that projects high-dimensional data into a lower-dimensional space.
We randomly select 1000 users in both domains for the recommendation scenarios \textbf{Phone\&Sport} and  \textbf{Sport\&Cloth}. 
The visualization results are shown in Figure \ref{disen_c_s} and Figure \ref{sport_cloth_disen_c_s}.
Furthermore, to assess whether the proposed model has effectively acquired domain-common knowledge from both source and target domains through shared users, 
we visualize the obfuscated domain-common features in both domains in Figure \ref{disen_source_target_c} and Figure \ref{sport_cloth_disen_source_target_c}.

We can observe a distinct separation between the domain-common embeddings and the domain-specific embeddings in Figure \ref{disen_c_s} and Figure \ref{sport_cloth_disen_c_s}.
This clear separation validates that our model can effectively disentangle user representations, ensuring that these obfuscated disentangled embeddings do not contain redundant information.
In Figure \ref{disen_source_target_c} and Figure \ref{sport_cloth_disen_source_target_c}, it is evident that these features are very similar, demonstrating that the obfuscated domain-common embeddings contain the shared information between these two domains.

 \subsection{Impact of Hyper-parameters (RQ4)}
\begin{figure}[!t]
\centering
\subfloat[]{\includegraphics[width=1.6in]{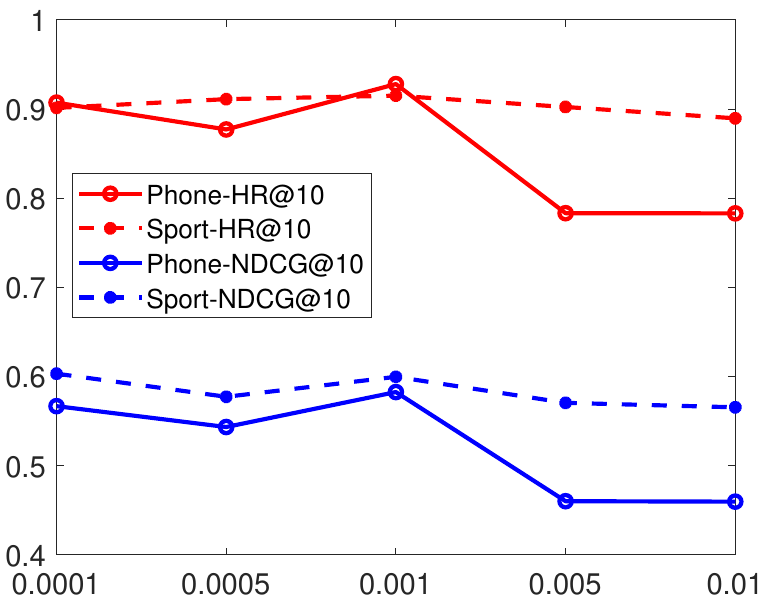}%
}
\hfil
\subfloat[]{\includegraphics[width=1.6in]{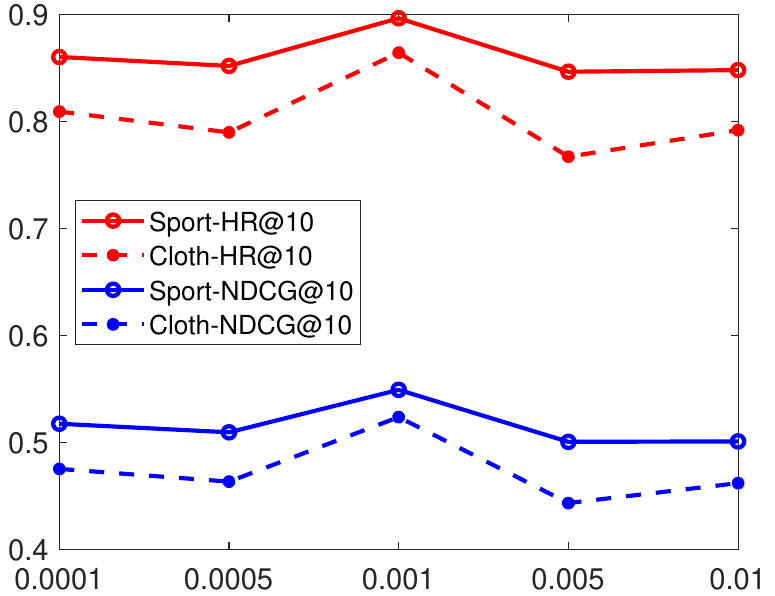}%
}
\hfil
\caption{Effect of different weight parameter $\alpha$ in scenarios: (a) \textbf{Phone\&Sport}  (b) \textbf{Sport\&Cloth}.}
\label{alpha}
\end{figure}

\begin{figure}[!t]
\centering
\subfloat[]{\includegraphics[width=1.6in]{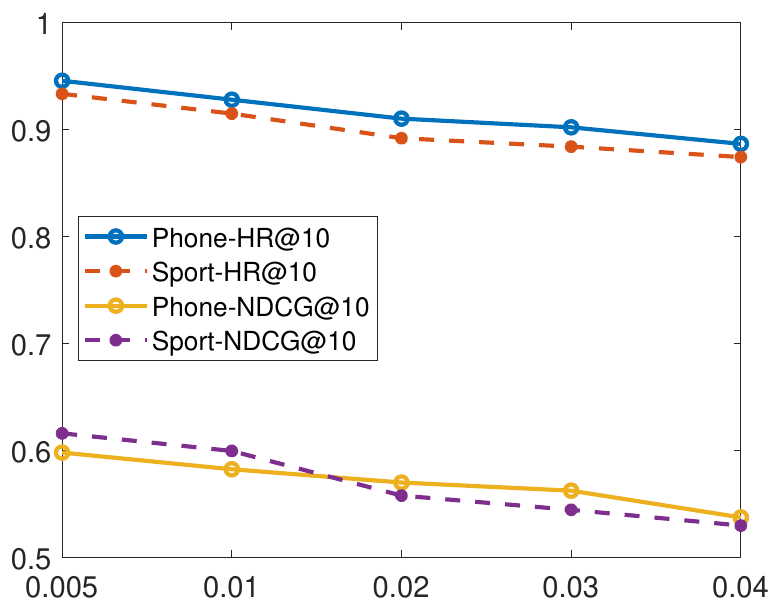}%
\label{lambda_phone_sport}}
\hfil
\subfloat[]{\includegraphics[width=1.6in]{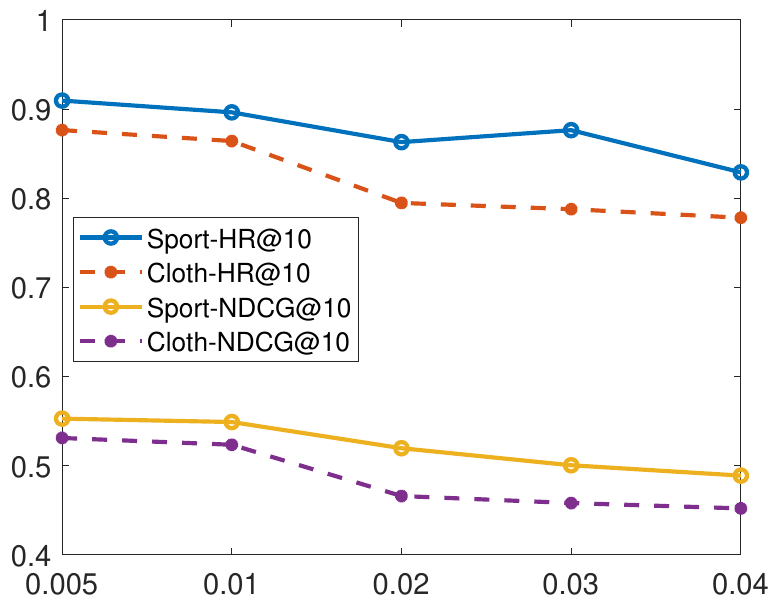}%
\label{lambda_sport_cloth}}
\hfil
\caption{Effect of different standard deviation $\lambda$ in scenarios: (a) \textbf{Phone\&Sport}  (b) \textbf{Sport\&Cloth}.}
\label{lambda}
\end{figure}

\begin{figure}[!t]
\centering
\subfloat[]{\includegraphics[width=1.6in]{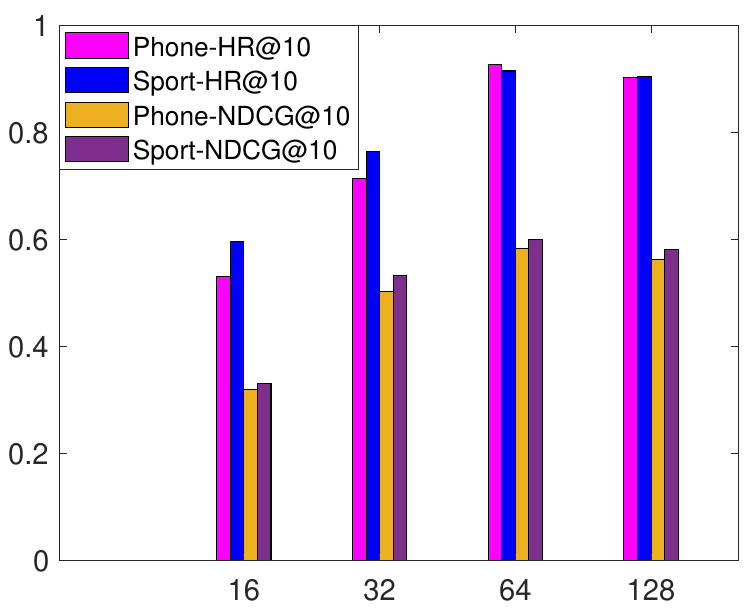}%
\label{emb_dim_phone_sport}}
\hfil
\subfloat[]{\includegraphics[width=1.6in]{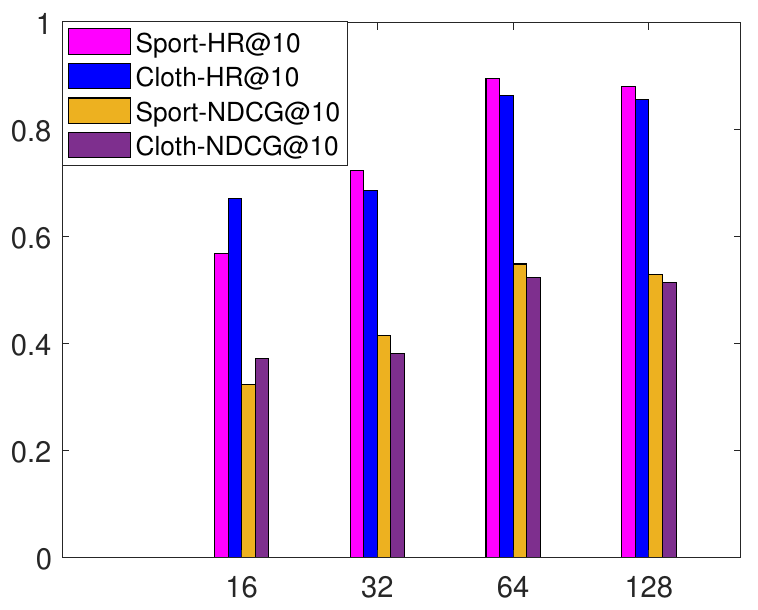}%
\label{emb_dim_sport_cloth}}
\hfil
\caption{Effect of different decoupled embedding dimensions in scenarios: (a) \textbf{Phone\&Sport}  (b) \textbf{Sport\&Cloth}.}
\label{emb_dim}
\end{figure}

In this section, we evaluate the model's performance under various settings of three crucial parameters: the weight parameter $\alpha$ for the contrastive learning loss, 
the standard deviation $\lambda$ for Laplace noise and the decoupling embedding dimension.
The comparative results in CDR scenarios \textbf{Phone\&Sport} and \textbf{Sport\&Cloth} are illustrated in Figure \ref{alpha}, Figure \ref{lambda} and Figure \ref{emb_dim}.

\subsubsection{Impact of $\alpha$}
P2M2-CDR achieves optimal results when $\alpha$ is set to 0.001. This suggests that a moderate weight parameter strikes a balance between aligning and separating disentangled embeddings while preserving overall recommendation performance.
\subsubsection{Impact of $\lambda$}
Regarding the standard deviation parameter $\lambda$, it is evident that a larger $\lambda$ introduces more noise, thereby enhancing user privacy protection. 
However, this increase in noise comes at the cost of a decrease in the model's performance.
We set $\lambda$ to 0.01, striking a balance between privacy preservation and maintaining acceptable performance levels.
 \subsubsection{Impact of decoupled embedding dimension}
We observe that the optimal performance is attained when the embedding dimension is set to 64.
Increasing the embedding dimension improves the model's effectiveness.
However, it is important to note that too large embedding dimensions may lead to overfitting.
Thus, choosing an appropriate embedding dimension is crucial to balance model complexity and performance.

 \subsection{Impact of various information fusion methods (RQ5)}
\begin{figure}[!t]
\centering
\subfloat[]{\includegraphics[width=1.6in]{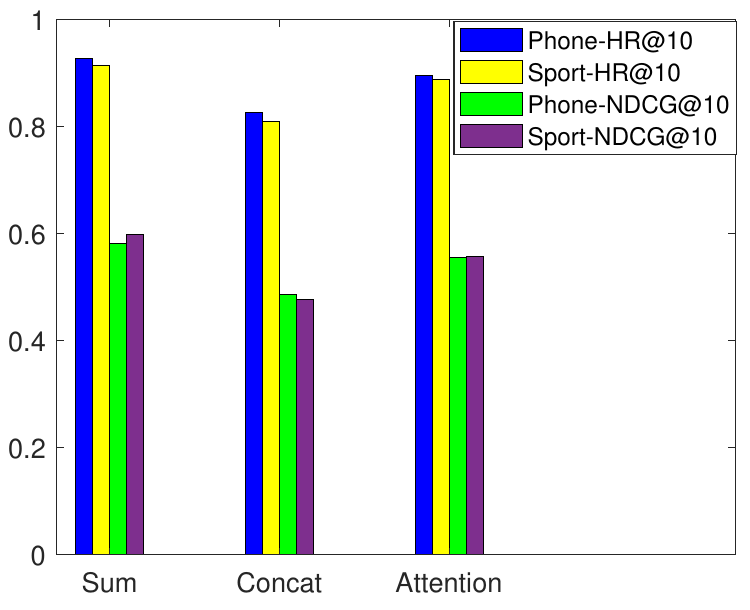}%
}
\hfil
\subfloat[]{\includegraphics[width=1.6in]{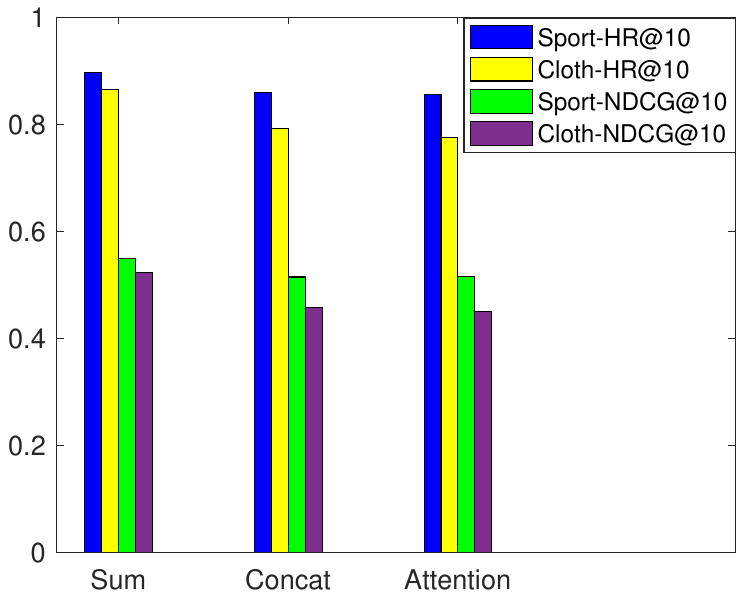}%
}
\hfil
\caption{Effect of different information fusion methods in scenarios: (a) \textbf{Phone\&Sport}  (b) \textbf{Sport\&Cloth}.}
\label{agg_way}
\end{figure}

 We employ three distinct aggregation methods, namely element-wise sum, concatenation, and attention, to combine obfuscated domain-common and domain-specific embeddings into the final user preferences.
 As illustrated in Figure \ref{agg_way}, it is evident that element-wise sum yields the highest performance, surpassing the other two methods by an average of 10.26\% and 2.89\% in terms of HR@10 within the Phone\&Sport scenario.
This superiority may be attributed to the fact that the element-wise sum method typically possesses fewer parameters and lower model complexity, thereby mitigating overfitting.

\section{Conclusion and Future Work}
In this work, we propose a privacy-preserving framework with multi-modal data for cross-domain recommendation (P2M2-CDR).
It contains a multi-modal disentangled encoder and a privacy-preserving decoder.
The multi-modal encoder integrates multi-modal features to learn comprehensive user and item representations and disentangles user representations into more informative domain-common and domain-specific embeddings.
The privacy-preserving decoder aims to introduce local differential privacy to protect user privacy when transferring knowledge across domains.
Experimental results on four CDR scenarios  Phone\&Elec, Phone\&Sport, Sport\&Cloth, and Elec\&Cloth demonstrate the effectiveness of our proposed model.

In the future, we intend to explore other prevalent privacy-preserving techniques, such as federated learning, 
to enhance model performance and simultaneously protect user privacy in cross-domain recommendation methods.
\\

\noindent \textbf{CRediT authorship contribution statement}\\

\textbf{Li Wang}: Conceptualization, Investigation, Methodology,
Software, Writing - original draft, Writing - review \& editing.
\textbf{Lei Sang}: Writing - review \& editing.
\textbf{Quangui Zhang}: Writing - review \& editing.
\textbf{Qiang Wu}: Supervision, Writing - review \& editing.
\textbf{Min Xu}: Supervision, Writing - review \& editing.
 \bibliographystyle{elsarticle-num} 
 \bibliography{references}





\end{document}